\documentclass[sigconf]{acmart}

\makeatletter                   
\def\mdseries@tt{m}             
\makeatother                    

\usepackage[draft=true]{minted} 
\usepackage{color}
\usepackage{hyperref}           
\hypersetup{
    colorlinks=true,
    linkcolor=blue,
    filecolor=red,      
    urlcolor=magenta,
    breaklinks=true,            
}

\usepackage{xcolor}

\renewcommand\footnotetextcopyrightpermission[1]{}

\begin{document}
\sloppy                         

\title{Language Modelling for Source Code with Transformer-XL}

\author{Thomas Dowdell}
\affiliation{%
  \institution{University of Newcastle}
  \city{Callaghan, NSW}
  \state{Australia}}
\email{tomjamesdowdell@gmail.com}
\author{Hongyu Zhang}
\affiliation{%
  \institution{University of Newcastle}
  \city{Callaghan, NSW}
  \state{Australia}}
\email{hongyu.zhang@newcastle.edu.au}

\begin{abstract}
It has been found that software, like natural language texts, exhibits "naturalness", which can be captured by statistical language models. In recent years, neural language models have been proposed to represent the naturalness of software through deep learning. In this paper, we conduct an experimental evaluation of state-of-the-art neural language models for source code, including RNN-based models and Transformer-XL based models. Through experiments on a large-scale Python code corpus, we find that the Transformer-XL model outperforms RNN-based models (including LSTM and GRU models) in capturing the naturalness of software, with far less computational cost.

\end{abstract}

\maketitle
\pagestyle{plain} 

\section{Introduction}

Over years of software development, a large amount of source code has been accumulated online. Through large-scale mining, it has been found that source code contains many repetitive, statistical regularities, which are also termed as "naturalness" of source code \cite{Hindle2012}\cite{learningnatural}\cite{allamanis2018survey}\cite{miningscale}\cite{8812062}. The naturalness hypothesis has been confirmed by empirical evidence \cite{bhoopchand2016}\cite{Cummins2017}\cite{Gupta2017}\cite{abs-1903-05734}\cite{sridhara2010towards}. Researchers have also applied natural language processing techniques, such as Language Modeling, to model the naturalness of source code  \cite{Hindle2012}\cite{abs-1903-05734} \cite{White2015} \cite{yang2019}. The constructed language models can be used in many practical programming tasks such as code completion \cite{2594321}\cite{Hindle2012}, syntax error fixing \cite{Umair2018}\cite{bhatia2016}\cite{8330219}, and API recommendation \cite{2950334}.

Language Models (LMs) are probability distributions over strings. Given a sequence of tokens $x_T$, language modelling is defined as predicting the token $x_t$ given the previous tokens from $x_1$ to $x_{t-1}$, i.e. $max_\theta P(x_t | x_{t-1}, x_{t-2}, ..., x_{t-1}, \theta)$, where $\theta$ are the trainable model variables. Multiple language models for source code have already been introduced \cite{Gupta2017} \cite{Mesbah2019} \cite{localsoftware}. A classical language modeling method is the N-gram model. Recent research uses RNN-based models (including LSTM and GRU models) \cite{2950334}\cite{Gupta2017}\cite{2594321}\cite{8330219} and find that RNN-based models superseded N-gram models.

Transformer models \cite{vaswani2017} have been recently introduced, and have outperformed RNNs for natural language processing tasks, because of their ability to track long-range dependencies and their parallelizable nature. Many Transformer variants have been introduced, notably the Transformer-XL\cite{dai2019}, which is capable of tracking remarkably long-range dependencies.

In this paper, we investigate the effectiveness of the Transformer-XL based language model for source code, where the model is trained to predict the next token, given a sequence of source code tokens.
Our experimental results on a Python dataset show that the Transformer-XL models largely outperform the state-of-the-art RNN-based models. This strongly suggests that the long-range analysis of the Transformer-XL is helpful for modelling software naturalness. The Transformer-XL model also contained significantly less time to train in comparison to both the LSTM and GRU models.

\section{Transformer-based Language Models for Source Code}

\subsection{The Transformer Model}

Although LSTM \cite{hochreiter1997} and GRU \cite{cho2014} models are considered the pinnacle of RNN-based models, both have noted problems with long-term dependencies. The Transformer model \cite{vaswani2017} has been introduced to overcome the limitations of RNN-based models. Transformer models do not calculate the results sequentially, but instead calculate the results in a parallel manner using a self-attention mechanism, also allowing a Transformer-based model to be trained in a shorter amount of time.  Transformers can achieve better results with less computational cost than the RNN-based models on a variety of natural language tasks \cite{delvin2018}. 

\begin{figure}[!h]
\vspace{-12pt}
\includegraphics[clip,trim={12cm 1.7cm 13cm 4.2cm},width=7cm]{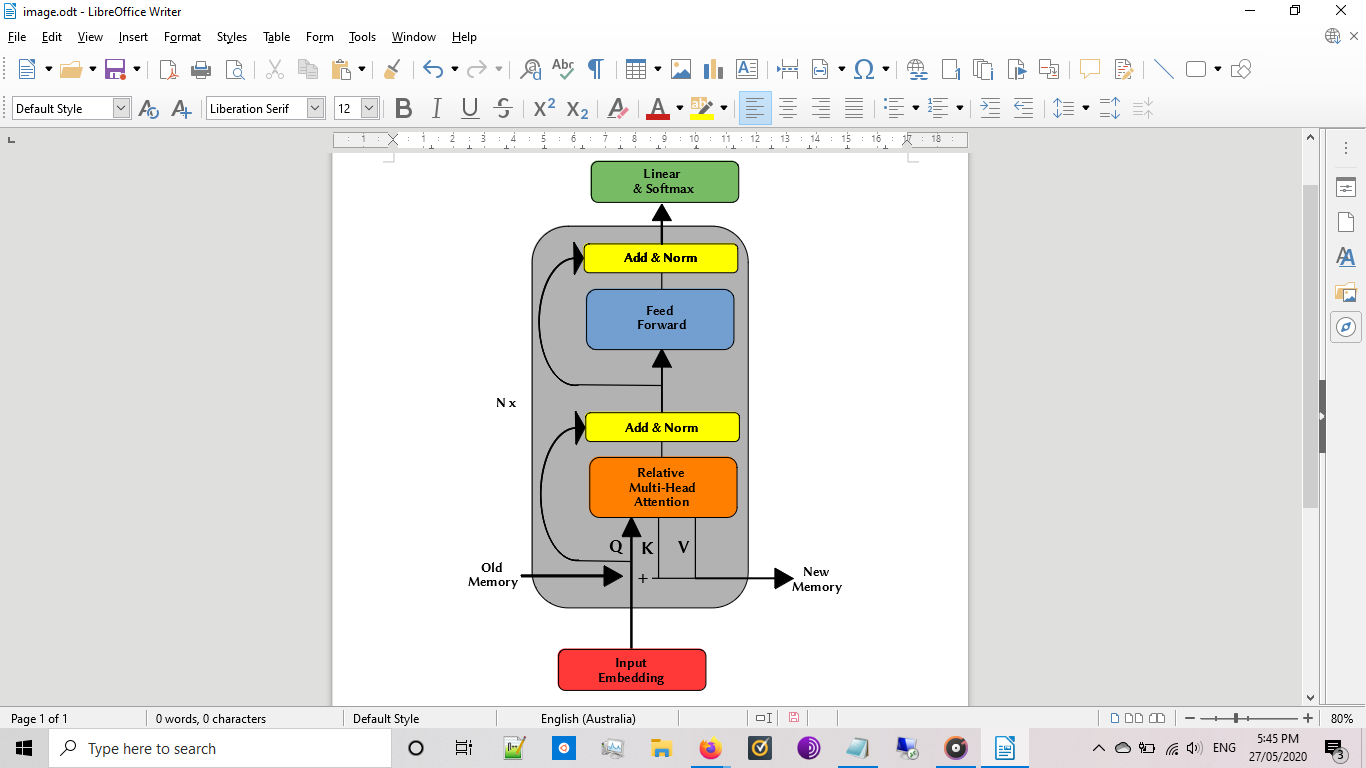}
\vspace{-6pt}
\caption{The Transformer-XL model, as specified in \cite{yang2019}.}
\label{fig:arch}
\vspace{-6pt}
\end{figure}

The key to the Transformer’s success is an attention block, defined as the self-attention mechanism, and a feedforward (FFD) block, both of which are calculated in parallel. The attention block allows a model to view all data across an entire sequence, unimpeded, in a computationally efficient manner, while the feedforward block analyzes the data in a sequence-independent manner. Transformer-based neural networks take less time to train than the RNNs, require less memory and achieve a lower per-token loss \cite{vaswani2017}. A single layer of the Transformer can be defined, mathematically, for the \textit{l}\textsuperscript{th} layer, as:

\[x_0 = inputs \]
\[att_l = Self-Attention\_Mechanism(x_{l - 1}) \]
\[x_l = AddNorm(att_l, x_{l - 1}) \]
\[ffd_l = FFD(x_l) \]
\[x_l = AddNorm(ffd_l, x_l) \]

The input sequence of tokens is defined as \textit{inputs}. The functions \textit{AddNorm}, \textit{FFD} and \textit{Self-Attention\_Mechanism} are defined as:

\[AddNorm(x, y) = layernorm(x) + y \]
\[FFD(x_l) = W_{l,2} * act(W_{l,1} * x_l + b_{l,1}) + b_{l,2} \]
\[Self-Attention\_Mechanism(x_l) = W_O * softmax(Q * K^T / \sqrt d_k) * V \]
\[Q, K, V = W_Q * x_l, W_K * x_l, W_V * x_l \]

The variables \textit{W\textsubscript{l,1}, W\textsubscript{l,2}, W\textsubscript{Q}, W\textsubscript{K}, W\textsubscript{V}} are all trainable weight matrices, the variables \textit{b\textsubscript{l,1} and b\textsubscript{l,2}} are trainable bias vectors and function \textit{act} is an user-defined non-linear activation function. In all models used in this paper, the GELU activation function \cite{gelu} is used.

The softmax function is applied over the output to yield a categorical probability distribution over each token. At the time-step \textit{t}, the Transformer calculates the input token for time-step \textit{t + 1}.

\subsection{The Transformer-XL Model} ~

The Transformer-XL \cite{dai2019} \cite{yang2019} analyzes data in an identical manner to a vanilla transformer, except that, for each sequence, the results calculated for each ‘layer’ of the transformer are saved and re-inputted, in a gradient-free manner, for the next sequence. This allows the Transformer-XL to ‘see’ across previous sequences, allowing for a greater amount of input information and therefore greater expressiveness and accuracy. This, conceptually, makes sense, given that a model with greater memory would be capable of analysing this larger memory to generate a better understanding of the underlying source code.

Formally, Transformer-XL redefines the equation that generates the values \textit{Q, K, V} of the attention mechanism as follows:

\[Q = W_Qx \]
\[K, V = W_{K,V}[\textit{SG}(x_{t-1}) * x] \]

\noindent where \textit{[ * ]} refers to concatenation, \textit{SG} refers to the stop-gradient function and \textit{x\textsubscript{t-1}} refers to the input to the attention mechanism from the previous sequence.

Figure 1 shows the structure of Transformer-XL model, as specified in \cite{yang2019}. With the exception of the memory-specific self-attention mechanism, the Transformer-XL layer is identical to the Transformer layer. The self-attention is normalized using layer normalization and a residual connection, followed by an FFD network, as specified in Section 2.1 above.

\section{Experiments}

\subsection{Data Collection and Preprocessing}

To perform the evaluation, we create a dataset composed of Python source code collected from Github. The dataset contains over 17,000 Python files, which total over 3 million individual lines of code. We use two different forms of tokenization; character-level tokenization and subword-level tokenization. Character-level tokenization contains 141M tokens, while subword tokenization contains 88M tokens.

We did not use word-level tokenization for this task. This is because, unlike natural language, source code does not have an explicit vocabulary. Therefore, there is no way to accurately and effectively model all possible words inside a source code file. However, the source code can be effectively modelled using characters and subwords\cite{sennrich2015}. 
For example, the source code \textit{print(x + 3 if x == 0)} can be tokenized into a sequence of subwords, [\textit{print, (, x, +, 3, if, x, ==,  0, )}].

We explicitly set, for subword tokenization, the vocabulary size hyperparameter to 1000.
We chose this vocabulary size because, as previous authors have noted \cite{lan2019}, a large vocabulary is responsible for one of the most computationally expensive aspects of the model. By setting the vocabulary size to 1000, the model can tokenize the most common words as a single token, but break down less common words into a series of subwords. This allows the model to express any input information without unacceptable computational complexity and information loss. Therefore, subword-level tokenization does not suffer from the out-of-vocabulary (OOV) problem.

Given this dataset, the model is trained to functionally predict the next token in the line, without being able to ‘peek' forward through the line. We specify that the model is trained on a sequence that includes 256 tokens, and the memory from the previous sequence.

We split the collected data in the training, validation and testing data using an 80\%/10\%/10\% split. We were careful to avoid as much code duplication as possible, which has been previously shown to be a common problem for many source code-based machine learning models \cite{allamanis2019}. In order to do this, we tested that each file in the training dataset did not reappear in either the validation or testing dataset. Further, for each validation and test file, we tested to make sure that the number of lines in each file was not repeated in the training dataset above a certain threshold. If the file was over the threshold, then it was removed from the dataset. We chose 25\% as the threshold.


\subsection{Model Training}

Each model is trained for 50 epochs, where each epoch contains 512 iterations. The learning rate is set to a linear-warmup from 1e-6 to 5e-4 for the first 5120 iterations, and a cosine-decay rate back to 1e-6 for the following iterations. The optimizer used is the Adam Optimizer \cite{kingma2014adam}. We clip the gradient norm to 0.1, which we found was essential for training. We noted that if the gradients were not clipped, then all models would produce inferior results.

Both RNN models and the Transformer-XL model have a hidden size of 512 units, and have a dropout rate set to 0.1 for training. Initially, we trained each model with a depth of 4 layers. RNN-based models do not typically include many more layers due to the RNN's massive computational cost. However, Transformer-based models typically have many more layers, anywhere from 12 \cite{vaswani2017} to 72 \cite{shoeybi2019} layers.

The depth of a model is considered an essential aspect of a model's ability to learn, but we noted that increasing the number of RNN layers significantly slows down training, but this was not the case for the Transformer-XL model. Considering previous papers\cite{alrfou} have shown that Transformer-based models achieve noticeably superior results by increasing the depth for language modelling, we decided to further experiment by increasing the number of layers to 8.


The experiments were primarily conducted on a single server and a single K40 NVIDIA Tesla GPU, alongside 128GB of RAM.

\subsection{Evaluation Metrics}
The evaluation metric used for character-level experiments in this paper is the BPC (bytes-per-character) metric \cite{alrfou}, where the BPC is calculated per-token. This metric is chosen because BPC is a function of test-entropy, which effectively displays how ‘confident' a model is about guessing the correct answer. The lower the BPC the more confident the model is when guessing the correct answer and the more effective the overall model. The cross-entropy equation can be defined as, given, at time-step $t$, the predicted output \textit{y\textsubscript{t}} and the target output \textit{z\textsubscript{t}}:

\[loss_t = -1 * z_t * log(y_t) \]

The BPC is defined as:

\[bpc = \frac{loss}{log(2)} \]

For subword-level experiments, the metric is the perplexity rather than the BPC. Perplexity is defined as:

\[perplexity = e\textsuperscript{loss} \]

Low perplexity is desirable since the perplexity is the exponential of the entropy. Multiple previous papers have stated that, for character-level analysis, BPC is the metric of choice but perplexity is the metric to be used for both word-level and subword-level analysis \cite{alrfou}\cite{delvin2018}\cite{yang2019}.

\subsection{Experimental Results}

The results from all experiments are presented in Table 1. 
We tested each model 3 times and presented the average results of each model.
Each model was initialized randomly, and the random-seed is different for each experiment. We noted that each model, when tested with the same data, performed similarly across different runs and seeds. The validation data loss, for both the character-level and BPC-level analysis, can be seen in Figure 2 and Figure 3.

\begin{table}[tb]
\begin{center}
    \begin{tabular}{|l|c|c|}
    \hline
    \textbf{Model} & \textbf{Character-Level} &  \textbf{Subword-Level}\\
    & \textbf{Test BPC} & \textbf{Test Perplexity} \\ \hline
    \textbf{LSTM (4 Layer)} & 1.8706 & 6.4552 \\
    & \textit{(0.0053)} & \textit{(0.0774)} \\
    \hline
    \textbf{GRU (4 Layer)} & 1.2758 & 6.1135 \\
    & \textit{(0.002)} & \textit{(0.0146)} \\
    \hline
    \textbf{Transformer-XL} & 1.1506 & 2.7278 \\
    \textbf{(4 Layer)} & \textit{(0.0018)} & \textit{(0.0005)} \\
    \hline
    \textbf{Transformer-XL} & \textbf{1.1297} & \textbf{2.7185} \\
    \textbf{(8 Layer)} & \textit{(0.0063)} & \textit{(0.0208)} \\
    \hline
    \end{tabular}
\end{center}
\caption{The test BPC and Perplexity of each of the three models, where better models output lower values. The numbers in brackets are standard deviation across the multiple runs. Across both tokenization schemes, the Transformer-XL model outperforms both RNN models.}
\vspace{-12pt}
\end{table}

\begin{figure}[!h]
\vspace{-12pt}
\includegraphics[clip,width=8cm]{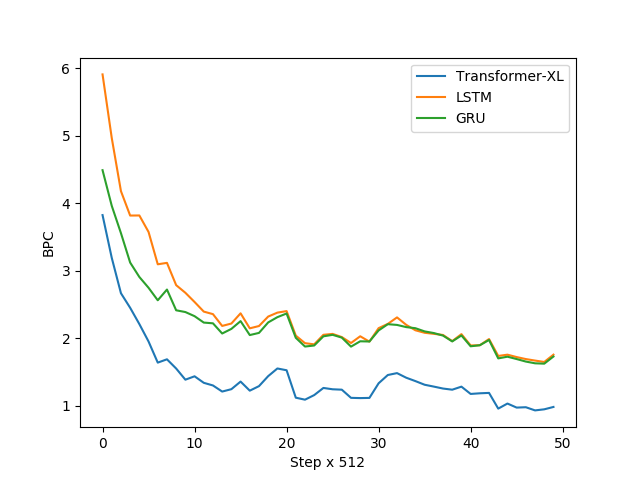}
\vspace{-6pt}
\caption{The BPC for the LSTM, GRU and Transformer-XL model when analysing character-level source code. The Transformer-XL, across the entire run, outperforms the RNN models. We chose to display the best-performing model for the LSTM, GRU and Transformer-XL.} 
\label{fig:arch}
\vspace{-6pt}
\end{figure}

The LSTM model, which has been used as the previous neural source code modelling work \cite{dam2016}, achieved the highest per-character BPC (1.8706) and per-subword Perplexity (6.4552). The GRU model achieved reliably better results than the LSTM (BPC 1.2758 and Perplexity 6.1135), but only by a small margin. The Transformer-XL model, on the other hand, could reliably outperform both the LSTM and GRU models by a substantial margin, for both the 4-layer and 8-layer models. 
The 8-layer Transformer-XL model, which performed better then all others, achieved a per-character BPC of 1.1297 and per-subword perplexity of 2.7185, which is lower than all other models. It is worth noting that the 4-layer Transformer-XL outperformed both RNN models as well. 

\begin{figure}[!h]
\vspace{-12pt}
\includegraphics[clip,width=8cm]{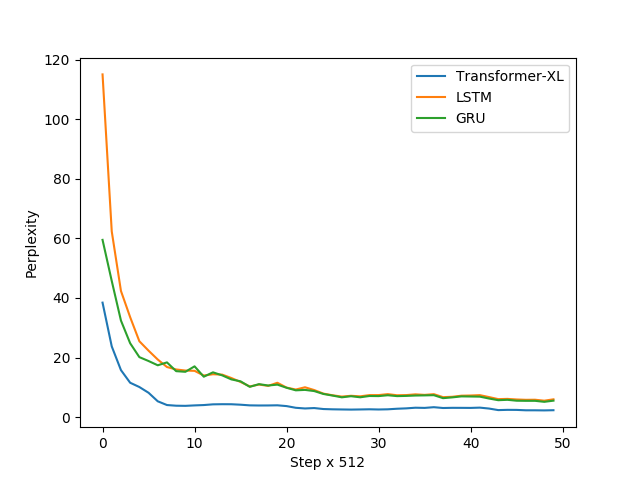}
\vspace{-6pt}
\caption{The Perplexity for the LSTM, GRU and Transformer-XL model when analysing subword-level source code. The Transformer-XL outperforms the RNN models.} 
\label{fig:arch}
\vspace{-6pt}
\end{figure}

The superior results of the 4-layer Transformer-XL model, in comparison to the RNN models, is likely a sign that the Transformer-XL model can extract more rich and useful features than both RNN models.

We also found that an 8-layer Transformer-XL model outperformed a 4-layer Transformer-XL model, but not by the same margin that a Transformer-XL outperforms the RNN. This 8-layer Transformer-XL, despite having twice the depth of the RNN models, was trained in less than half of the time required by the RNN models, even when trained on identical hardware. 

We note that the training time for each model depends on the hardware that the model is trained on. For example, an RNN model that is trained on a quicker GPU will be trained faster than a more efficient model on a more slow GPU. In order to reflect this, we recorded the normalized training time, where each model was trained on the same GPU and the time to train was normalized, to better reflect the length of training for each model in comparison to the other models. The normalized training time for each model can be seen in Table 2, from which we can see that Transformer-XL models require much less time to train than the two RNN models (LSTM and GRU).

In summary, our experiments show that the Transformer-XL model outperformed both RNN models when the layers are all set to 4, and the Transformer-XL further improves results when there are 8 layers instead of 4. Furthermore, even though the 8-layer Transformer-XL model has twice the depth and more trainable parameters, it takes less than half of the time to train and achieves notably superior results. 

\section{Discussion}
Our experimental results
suggest that the Transformer-XL model is a substantially superior model to either the LSTM or GRU model for modelling source code. This is because the RNN models cannot easily be trained for more layers without incurring massive computational cost, and therefore the RNN models produce inferior result \cite{vaswani2017}.
The Transformer-XL model, on the other hand, can produce noticeably superior results by increasing the layers without an unacceptable computational cost. Therefore, the Transformer-XL can achieve substantially superior using results in practice.  
We suggest using Transformer-XL in future research work on language modelling for source code.

Previous work on natural language processing has shown that models that are trained on language modelling based tasks can achieve state-of-the-art results on downstream NLP tasks, such as sentiment analysis and question-answering \cite{delvin2018}.
We would hypothesize that downstream source code related tasks can be performed in a similar way. 
Pre-training the models to perform language modelling allows a model to ``learn" the underlying statistical behaviour (naturalness) of source code, and this understanding can be leveraged to effectively perform downstream tasks.

Future work could focus on applying language models 
to the source code related downstream tasks such as automatic code completion. 
These models could also be pushed further. For example, these models could be easily retrained to perform more complicated tasks such as automatic error detection or automatic bug fixing. These tasks cannot be expressed in an identical manner to language modelling, in a manner similar to automatic code completion. However, 
the software naturalness
that the model learns can be leveraged to better perform these downstream tasks. Future work should focus on testing the effectiveness of these downstream tasks, and whether a language model can be trained to improve these tasks effectively.

\begin{table}[tb]
\begin{center}
    \begin{tabular}{|l|c|}
    \hline
    \textbf{Model} & \textbf{Normalized} \\
    & \textbf{Training Time} \\ \hline
    \textbf{LSTM (4 Layer)} & 4.2 \\ 
    \hline
    \textbf{GRU (4 Layer)} & 3.58 \\ 
    \hline
    \textbf{Transformer-XL} & 1 \\ 
    \textbf{(4 Layer)} & \\
    \hline
    \textbf{Transformer-XL} & 1.39 \\
    \textbf{(8 Layer)} & \\
    \hline
    \end{tabular}
\end{center}
\caption{The normalized training time for each model, where each unit is the normalized length of time for training a model over a single iteration. The 4-layer Transformer-XL model requires the shortest training time, followed by the 8-layer Transformer-XL model. It is worth noting that the 8-layer Transformer-XL, despite having twice the depth of RNNs, takes less than half of the time to train.} 
\vspace{-12pt}
\end{table}

\section{Conclusion}

This paper has shown that Transformer-based language models largely outperform RNN-based models for source code modelling. The Transformer-XL model achieves substantially better results with less than half the complexity and training time.

Based on this work, we recommend that Transformer-XL can be adopted for language modelling tasks relating to source code. 
Following the examples set in natural language processing \cite{delvin2018}, this model can be further applied to other downstream source code related tasks, such as automatic code completion and automatic bug fixing. 

Our tool and experimental data are publicly available at \url{https://
github.com/Anon-111/Source-Code-Modelling}. 

\bibliographystyle{ACM-Reference-Format}
\bibliography{acmart}


\begin{thebibliography}{32}


\ifx \showCODEN    \undefined \def \showCODEN     #1{\unskip}     \fi
\ifx \showDOI      \undefined \def \showDOI       #1{#1}\fi
\ifx \showISBNx    \undefined \def \showISBNx     #1{\unskip}     \fi
\ifx \showISBNxiii \undefined \def \showISBNxiii  #1{\unskip}     \fi
\ifx \showISSN     \undefined \def \showISSN      #1{\unskip}     \fi
\ifx \showLCCN     \undefined \def \showLCCN      #1{\unskip}     \fi
\ifx \shownote     \undefined \def \shownote      #1{#1}          \fi
\ifx \showarticletitle \undefined \def \showarticletitle #1{#1}   \fi
\ifx \showURL      \undefined \def \showURL       {\relax}        \fi
\providecommand\bibfield[2]{#2}
\providecommand\bibinfo[2]{#2}
\providecommand\natexlab[1]{#1}
\providecommand\showeprint[2][]{arXiv:#2}

\bibitem[\protect\citeauthoryear{Ahmed, Kumar, Karkare, Kar, and Gulwani}{Ahmed
  et~al\mbox{.}}{2018}]%
        {Umair2018}
\bibfield{author}{\bibinfo{person}{Umair Ahmed}, \bibinfo{person}{Pawan Kumar},
  \bibinfo{person}{Amey Karkare}, \bibinfo{person}{Purushottam Kar}, {and}
  \bibinfo{person}{Sumit Gulwani}.} \bibinfo{year}{2018}\natexlab{}.
\newblock \showarticletitle{Compilation of Error Repair: For the Student
  Programs, from the Student Programs}. In \bibinfo{booktitle}{\emph{40th
  International Conference on Software Engineering: Software Engineering
  Education and Training Track}} \emph{(\bibinfo{series}{ICSE-SEET’18})}.
  \bibinfo{publisher}{ACM}, \bibinfo{pages}{78--87}.
\newblock
\urldef\tempurl%
\url{https://dl.acm.org/citation.cfm?doid=3183377.3183383}
\showURL{%
\tempurl}


\bibitem[\protect\citeauthoryear{Al-Rfou, Choe, Constant, Guo, and
  Jones}{Al-Rfou et~al\mbox{.}}{2018}]%
        {alrfou}
\bibfield{author}{\bibinfo{person}{Rami Al-Rfou}, \bibinfo{person}{Dokook
  Choe}, \bibinfo{person}{Noah Constant}, \bibinfo{person}{Mandy Guo}, {and}
  \bibinfo{person}{Llion Jones}.} \bibinfo{year}{2018}\natexlab{}.
\newblock \showarticletitle{Character-Level Language Modeling with Deeper
  Self-Attention}.
\newblock \bibinfo{journal}{\emph{arXiv preprint arXiv:1808.04444v2}}
  (\bibinfo{year}{2018}).
\newblock


\bibitem[\protect\citeauthoryear{Allamanis}{Allamanis}{2018}]%
        {allamanis2019}
\bibfield{author}{\bibinfo{person}{Miltiadis Allamanis}.}
  \bibinfo{year}{2018}\natexlab{}.
\newblock \showarticletitle{The Adverse Effects of Code Duplication in Machine
  Learning Models of Code}.
\newblock \bibinfo{journal}{\emph{arXiv preprint arXiv:1812.06468}}
  (\bibinfo{year}{2018}).
\newblock


\bibitem[\protect\citeauthoryear{Allamanis, Barr, Bird, and Sutton}{Allamanis
  et~al\mbox{.}}{2014}]%
        {learningnatural}
\bibfield{author}{\bibinfo{person}{Miltiadis Allamanis},
  \bibinfo{person}{Earl~T. Barr}, \bibinfo{person}{Christian Bird}, {and}
  \bibinfo{person}{Charles Sutton}.} \bibinfo{year}{2014}\natexlab{}.
\newblock \showarticletitle{Learning Natural Coding Conventions}.
\newblock \bibinfo{journal}{\emph{arXiv preprint arXiv:1402.4182}}
  (\bibinfo{year}{2014}).
\newblock


\bibitem[\protect\citeauthoryear{Allamanis, Barr, Devanbu, and
  Sutton}{Allamanis et~al\mbox{.}}{2018}]%
        {allamanis2018survey}
\bibfield{author}{\bibinfo{person}{Miltiadis Allamanis},
  \bibinfo{person}{Earl~T Barr}, \bibinfo{person}{Premkumar Devanbu}, {and}
  \bibinfo{person}{Charles Sutton}.} \bibinfo{year}{2018}\natexlab{}.
\newblock \showarticletitle{A survey of machine learning for big code and
  naturalness}.
\newblock \bibinfo{journal}{\emph{ACM Computing Surveys (CSUR)}}
  \bibinfo{volume}{51}, \bibinfo{number}{4} (\bibinfo{year}{2018}),
  \bibinfo{pages}{81}.
\newblock


\bibitem[\protect\citeauthoryear{Allamanis and Sutton}{Allamanis and
  Sutton}{2013}]%
        {miningscale}
\bibfield{author}{\bibinfo{person}{Miltiadis Allamanis} {and}
  \bibinfo{person}{Charles Sutton}.} \bibinfo{year}{2013}\natexlab{}.
\newblock \showarticletitle{Mining Source Code Repositories at Massive Scale
  using Language Modeling}. In \bibinfo{booktitle}{\emph{2013 10th Working
  Conference on Mining Software Repositories (MSR)}}. IEEE,
  \bibinfo{pages}{207--216}.
\newblock


\bibitem[\protect\citeauthoryear{Bhatia and Singh}{Bhatia and Singh}{2016}]%
        {bhatia2016}
\bibfield{author}{\bibinfo{person}{Sahil Bhatia} {and} \bibinfo{person}{Rishabh
  Singh}.} \bibinfo{year}{2016}\natexlab{}.
\newblock \showarticletitle{Automated Correction for Syntax Errors in
  Programming Assignments using Recurrent Neural Networks}.
\newblock \bibinfo{journal}{\emph{arXiv preprint arXiv:1603.06129}}
  (\bibinfo{year}{2016}).
\newblock


\bibitem[\protect\citeauthoryear{Bhoopchand, Rockt~aschel, Barr, and
  Riedel}{Bhoopchand et~al\mbox{.}}{2016}]%
        {bhoopchand2016}
\bibfield{author}{\bibinfo{person}{Avishkar Bhoopchand}, \bibinfo{person}{Tim
  Rockt~aschel}, \bibinfo{person}{Earl Barr}, {and} \bibinfo{person}{Sebastian
  Riedel}.} \bibinfo{year}{2016}\natexlab{}.
\newblock \showarticletitle{Learning Python Code Suggestion with a Sparse
  Pointer Network}.
\newblock \bibinfo{journal}{\emph{arXiv preprint arXiv:1611.08307}}
  (\bibinfo{year}{2016}).
\newblock


\bibitem[\protect\citeauthoryear{Cho, van Merrienboer, Gulcehre, Bahdanau,
  Bougares, Schwenk, and Bengio}{Cho et~al\mbox{.}}{2014}]%
        {cho2014}
\bibfield{author}{\bibinfo{person}{Kyunghyun Cho}, \bibinfo{person}{Bart van
  Merrienboer}, \bibinfo{person}{Caglar Gulcehre}, \bibinfo{person}{Dzmitry
  Bahdanau}, \bibinfo{person}{Fethi Bougares}, \bibinfo{person}{Holger
  Schwenk}, {and} \bibinfo{person}{Yoshua Bengio}.}
  \bibinfo{year}{2014}\natexlab{}.
\newblock \showarticletitle{Learning Phrase Representations using RNN
  Encoder-Decoder for Statistical Machine Translation}.
\newblock \bibinfo{journal}{\emph{arXiv preprint arXiv:1406.1078}}
  (\bibinfo{year}{2014}).
\newblock


\bibitem[\protect\citeauthoryear{Cummins, Petoumenos, Wang, and
  Leather}{Cummins et~al\mbox{.}}{2017}]%
        {Cummins2017}
\bibfield{author}{\bibinfo{person}{Chris Cummins}, \bibinfo{person}{Pavlos
  Petoumenos}, \bibinfo{person}{Zheng Wang}, {and} \bibinfo{person}{Hugh
  Leather}.} \bibinfo{year}{2017}\natexlab{}.
\newblock \showarticletitle{Synthesizing benchmarks for predictive modeling}.
  In \bibinfo{booktitle}{\emph{Proceedings of the 2017 International Symposium
  on Code Generation and Optimization}} \emph{(\bibinfo{series}{CGO '17})}.
  \bibinfo{publisher}{ACM}, \bibinfo{pages}{86--99}.
\newblock
\urldef\tempurl%
\url{3049843}
\showURL{%
\tempurl}


\bibitem[\protect\citeauthoryear{Dai, Yang, Yang, Carbonell, Le, and
  Salakhutdinov}{Dai et~al\mbox{.}}{2019}]%
        {dai2019}
\bibfield{author}{\bibinfo{person}{Zihing Dai}, \bibinfo{person}{Zhilin Yang},
  \bibinfo{person}{Yiming Yang}, \bibinfo{person}{Jaime Carbonell},
  \bibinfo{person}{Quoc~V. Le}, {and} \bibinfo{person}{Ruslan Salakhutdinov}.}
  \bibinfo{year}{2019}\natexlab{}.
\newblock \showarticletitle{Transformer-XL: Attentive Language Models Beyond a
  Fixed-Length Context}.
\newblock \bibinfo{journal}{\emph{arXiv preprint arXiv:1901.02860}}
  (\bibinfo{year}{2019}).
\newblock


\bibitem[\protect\citeauthoryear{Dam, Tran, and Pham}{Dam
  et~al\mbox{.}}{2016}]%
        {dam2016}
\bibfield{author}{\bibinfo{person}{Hoa~Khanh Dam}, \bibinfo{person}{Truyen
  Tran}, {and} \bibinfo{person}{Trang Pham}.} \bibinfo{year}{2016}\natexlab{}.
\newblock \showarticletitle{A Deep Language Model Model for Software Code}.
\newblock \bibinfo{journal}{\emph{arXiv preprint arXiv:1608.02715}}
  (\bibinfo{year}{2016}).
\newblock


\bibitem[\protect\citeauthoryear{Delvin, Chang, Lee, and Toutanova}{Delvin
  et~al\mbox{.}}{2018}]%
        {delvin2018}
\bibfield{author}{\bibinfo{person}{Jacob Delvin}, \bibinfo{person}{Ming-Wei
  Chang}, \bibinfo{person}{Kenton Lee}, {and} \bibinfo{person}{Kristina
  Toutanova}.} \bibinfo{year}{2018}\natexlab{}.
\newblock \showarticletitle{BERT: pre-training of deep bidirectional
  transformers for language understanding}.
\newblock \bibinfo{journal}{\emph{arXiv preprint arXiv:1810.04805}}
  (\bibinfo{year}{2018}).
\newblock


\bibitem[\protect\citeauthoryear{Gu, Zhang, Zhang, and Kim}{Gu
  et~al\mbox{.}}{2016}]%
        {2950334}
\bibfield{author}{\bibinfo{person}{Xiaodong Gu}, \bibinfo{person}{Hongyu
  Zhang}, \bibinfo{person}{Dongmei Zhang}, {and} \bibinfo{person}{Sunghun
  Kim}.} \bibinfo{year}{2016}\natexlab{}.
\newblock \showarticletitle{Deep API Learning}. In
  \bibinfo{booktitle}{\emph{Proceedings of the 2016 24th ACM SIGSOFT
  International Symposium on Foundations of Software Engineering}}
  \emph{(\bibinfo{series}{FSE 2016})}. \bibinfo{publisher}{Association for
  Computing Machinery}, \bibinfo{address}{New York, NY, USA},
  \bibinfo{pages}{631–642}.
\newblock
\showISBNx{9781450342186}
\urldef\tempurl%
\url{https://doi.org/10.1145/2950290.2950334}
\showDOI{\tempurl}


\bibitem[\protect\citeauthoryear{Gupta, Pal, Kanade, and Shevade}{Gupta
  et~al\mbox{.}}{2017}]%
        {Gupta2017}
\bibfield{author}{\bibinfo{person}{Rahul Gupta}, \bibinfo{person}{Soham Pal},
  \bibinfo{person}{Aditya Kanade}, {and} \bibinfo{person}{Shirish Shevade}.}
  \bibinfo{year}{2017}\natexlab{}.
\newblock \showarticletitle{DeepFix: Fixing Common C Language Errors by Deep
  Learning}. In \bibinfo{booktitle}{\emph{Proceedings of the Thirty-First AAAI
  Conference on Artificial Intelligence}} \emph{(\bibinfo{series}{AAAI'17})}.
  \bibinfo{publisher}{AAAI Press}, \bibinfo{pages}{1345--1351}.
\newblock
\urldef\tempurl%
\url{http://dl.acm.org/citation.cfm?id=3298239.3298436}
\showURL{%
\tempurl}


\bibitem[\protect\citeauthoryear{Hendrycks and Gimpe}{Hendrycks and
  Gimpe}{2018}]%
        {gelu}
\bibfield{author}{\bibinfo{person}{Dan Hendrycks} {and} \bibinfo{person}{Kevin
  Gimpe}.} \bibinfo{year}{2018}\natexlab{}.
\newblock \showarticletitle{Gaussian Error Linear Units (GELUs)}.
\newblock \bibinfo{journal}{\emph{arXiv preprint arXiv:arXiv:1606.08415v3}}
  (\bibinfo{year}{2018}).
\newblock


\bibitem[\protect\citeauthoryear{Hindle, Barr, Su, Gabel, and Devanbu}{Hindle
  et~al\mbox{.}}{2012}]%
        {Hindle2012}
\bibfield{author}{\bibinfo{person}{Abram Hindle}, \bibinfo{person}{Earl~T.
  Barr}, \bibinfo{person}{Zhendong Su}, \bibinfo{person}{Mark Gabel}, {and}
  \bibinfo{person}{Premkumar Devanbu}.} \bibinfo{year}{2012}\natexlab{}.
\newblock \showarticletitle{On the Naturalness of Software}. In
  \bibinfo{booktitle}{\emph{Proceedings of the 34th International Conference on
  Software Engineering}} \emph{(\bibinfo{series}{ICSE '12})}.
  \bibinfo{publisher}{IEEE Press}, \bibinfo{address}{Piscataway, NJ, USA},
  \bibinfo{pages}{837--847}.
\newblock
\showISBNx{978-1-4673-1067-3}
\urldef\tempurl%
\url{http://dl.acm.org/citation.cfm?id=2337223.2337322}
\showURL{%
\tempurl}


\bibitem[\protect\citeauthoryear{Hochreiter and Schmidhuber}{Hochreiter and
  Schmidhuber}{1997}]%
        {hochreiter1997}
\bibfield{author}{\bibinfo{person}{Sepp Hochreiter} {and}
  \bibinfo{person}{Jürgen Schmidhuber}.} \bibinfo{year}{1997}\natexlab{}.
\newblock \showarticletitle{Long Short-Term Memory}. In
  \bibinfo{booktitle}{\emph{Neural Computation}}. \bibinfo{pages}{1735--1780}.
\newblock
\urldef\tempurl%
\url{https://www.mitpressjournals.org/doi/10.1162/neco.1997.9.8.1735}
\showURL{%
\tempurl}


\bibitem[\protect\citeauthoryear{Karampatsis and Sutton}{Karampatsis and
  Sutton}{2019}]%
        {abs-1903-05734}
\bibfield{author}{\bibinfo{person}{Rafael{-}Michael Karampatsis} {and}
  \bibinfo{person}{Charles~A. Sutton}.} \bibinfo{year}{2019}\natexlab{}.
\newblock \showarticletitle{Maybe Deep Neural Networks are the Best Choice for
  Modeling Source Code}.
\newblock \bibinfo{journal}{\emph{CoRR}}  \bibinfo{volume}{abs/1903.05734}
  (\bibinfo{year}{2019}).
\newblock
\showeprint[arxiv]{1903.05734}
\urldef\tempurl%
\url{http://arxiv.org/abs/1903.05734}
\showURL{%
\tempurl}


\bibitem[\protect\citeauthoryear{Kingma and Ba}{Kingma and Ba}{2014}]%
        {kingma2014adam}
\bibfield{author}{\bibinfo{person}{Diederik~P Kingma} {and}
  \bibinfo{person}{Jimmy Ba}.} \bibinfo{year}{2014}\natexlab{}.
\newblock \showarticletitle{Adam: A method for stochastic optimization}.
\newblock \bibinfo{journal}{\emph{arXiv preprint arXiv:1412.6980}}
  (\bibinfo{year}{2014}).
\newblock


\bibitem[\protect\citeauthoryear{Lan, Chen, Goodman, Gimpel, Sharma, and
  Soricut}{Lan et~al\mbox{.}}{2019}]%
        {lan2019}
\bibfield{author}{\bibinfo{person}{Zhenzhong Lan}, \bibinfo{person}{Mingda
  Chen}, \bibinfo{person}{Sebastian Goodman}, \bibinfo{person}{Kevin Gimpel},
  \bibinfo{person}{Piyush Sharma}, {and} \bibinfo{person}{Radu Soricut}.}
  \bibinfo{year}{2019}\natexlab{}.
\newblock \showarticletitle{ALBERT: A Lite BERT for Self-supervised Learning of
  Language Representations}.
\newblock \bibinfo{journal}{\emph{arXiv preprint arXiv:1909.11942}}
  (\bibinfo{year}{2019}).
\newblock


\bibitem[\protect\citeauthoryear{Mesbah, Rice, Johnston, Glorioso, and
  Aftandilian}{Mesbah et~al\mbox{.}}{2019}]%
        {Mesbah2019}
\bibfield{author}{\bibinfo{person}{Ali Mesbah}, \bibinfo{person}{Andrew Rice},
  \bibinfo{person}{Emily Johnston}, \bibinfo{person}{Nick Glorioso}, {and}
  \bibinfo{person}{Edward Aftandilian}.} \bibinfo{year}{2019}\natexlab{}.
\newblock \showarticletitle{DeepDelta: learning to repair compilation errors}.
  In \bibinfo{booktitle}{\emph{Proceedings of the 2019 27th ACM Joint Meeting
  on European Software Engineering Conference and Symposium on the Foundations
  of Software Engineering}} \emph{(\bibinfo{series}{ESEC/FSE 2019})}.
  \bibinfo{publisher}{ACM}, \bibinfo{pages}{925--936}.
\newblock
\urldef\tempurl%
\url{https://dl.acm.org/citation.cfm?id=3340455}
\showURL{%
\tempurl}


\bibitem[\protect\citeauthoryear{Raychev, Vechev, and Yahav}{Raychev
  et~al\mbox{.}}{2014}]%
        {2594321}
\bibfield{author}{\bibinfo{person}{Veselin Raychev}, \bibinfo{person}{Martin
  Vechev}, {and} \bibinfo{person}{Eran Yahav}.}
  \bibinfo{year}{2014}\natexlab{}.
\newblock \showarticletitle{Code Completion with Statistical Language Models}.
\newblock \bibinfo{journal}{\emph{SIGPLAN Not.}} \bibinfo{volume}{49},
  \bibinfo{number}{6} (\bibinfo{date}{June} \bibinfo{year}{2014}),
  \bibinfo{pages}{419–428}.
\newblock
\showISSN{0362-1340}
\urldef\tempurl%
\url{https://doi.org/10.1145/2666356.2594321}
\showDOI{\tempurl}


\bibitem[\protect\citeauthoryear{{Santos}, {Campbell}, {Patel}, {Hindle}, and
  {Amaral}}{{Santos} et~al\mbox{.}}{2018}]%
        {8330219}
\bibfield{author}{\bibinfo{person}{E.~A. {Santos}}, \bibinfo{person}{J.~C.
  {Campbell}}, \bibinfo{person}{D. {Patel}}, \bibinfo{person}{A. {Hindle}},
  {and} \bibinfo{person}{J.~N. {Amaral}}.} \bibinfo{year}{2018}\natexlab{}.
\newblock \showarticletitle{Syntax and sensibility: Using language models to
  detect and correct syntax errors}. In \bibinfo{booktitle}{\emph{2018 IEEE
  25th International Conference on Software Analysis, Evolution and
  Reengineering (SANER)}}. \bibinfo{pages}{311--322}.
\newblock
\showISSN{null}
\urldef\tempurl%
\url{https://doi.org/10.1109/SANER.2018.8330219}
\showDOI{\tempurl}


\bibitem[\protect\citeauthoryear{Sennrich, Haddow, and Birch}{Sennrich
  et~al\mbox{.}}{2015}]%
        {sennrich2015}
\bibfield{author}{\bibinfo{person}{Rico Sennrich}, \bibinfo{person}{Barry
  Haddow}, {and} \bibinfo{person}{Alexandra Birch}.}
  \bibinfo{year}{2015}\natexlab{}.
\newblock \showarticletitle{Neural Machine Translation of Rare Words with
  Subword Units}.
\newblock \bibinfo{journal}{\emph{arXiv preprint arXiv:1509.07909}}
  (\bibinfo{year}{2015}).
\newblock


\bibitem[\protect\citeauthoryear{Shoeybi, Patwary, Puri, LeGresley, Casper, and
  Catanzaro}{Shoeybi et~al\mbox{.}}{2019}]%
        {shoeybi2019}
\bibfield{author}{\bibinfo{person}{Mohammad Shoeybi}, \bibinfo{person}{Mostofa
  Patwary}, \bibinfo{person}{Raul Puri}, \bibinfo{person}{Patrick LeGresley},
  \bibinfo{person}{Jard Casper}, {and} \bibinfo{person}{Bryan Catanzaro}.}
  \bibinfo{year}{2019}\natexlab{}.
\newblock \showarticletitle{Megatrom-LM: Training Multi-Billion Parameter
  Language Models Using Model Parallelism}.
\newblock \bibinfo{journal}{\emph{arXiv preprint arXiv:arXiv:1909.08053v3}}
  (\bibinfo{year}{2019}).
\newblock


\bibitem[\protect\citeauthoryear{Sridhara, Hill, Muppaneni, Pollock, and
  Vijay-Shanker}{Sridhara et~al\mbox{.}}{2010}]%
        {sridhara2010towards}
\bibfield{author}{\bibinfo{person}{Giriprasad Sridhara}, \bibinfo{person}{Emily
  Hill}, \bibinfo{person}{Divya Muppaneni}, \bibinfo{person}{Lori Pollock},
  {and} \bibinfo{person}{K Vijay-Shanker}.} \bibinfo{year}{2010}\natexlab{}.
\newblock \showarticletitle{Towards automatically generating summary comments
  for java methods}. In \bibinfo{booktitle}{\emph{Proceedings of the IEEE/ACM
  international conference on Automated software engineering}}. ACM,
  \bibinfo{pages}{43--52}.
\newblock


\bibitem[\protect\citeauthoryear{Tu, Su, and Devanbu}{Tu et~al\mbox{.}}{2014}]%
        {localsoftware}
\bibfield{author}{\bibinfo{person}{Zhaopeng Tu}, \bibinfo{person}{Zheopeng Su},
  {and} \bibinfo{person}{Premkumar Devanbu}.} \bibinfo{year}{2014}\natexlab{}.
\newblock \showarticletitle{On the localness of software}. In
  \bibinfo{booktitle}{\emph{Proceedings of the 22nd ACM SIGSOFT International
  Symposium on Foundations of Software Engineering}}.
  \bibinfo{pages}{269--280}.
\newblock


\bibitem[\protect\citeauthoryear{Vaswani, Sheezer, Parmar, Uszkoreit, Jones,
  Gomez, Kaiser, and Polosukhin}{Vaswani et~al\mbox{.}}{2017}]%
        {vaswani2017}
\bibfield{author}{\bibinfo{person}{Ashish Vaswani}, \bibinfo{person}{Noam
  Sheezer}, \bibinfo{person}{Niki Parmar}, \bibinfo{person}{Jakob Uszkoreit},
  \bibinfo{person}{Llion Jones}, \bibinfo{person}{Aidan Gomez},
  \bibinfo{person}{Lukasz Kaiser}, {and} \bibinfo{person}{Illia Polosukhin}.}
  \bibinfo{year}{2017}\natexlab{}.
\newblock \showarticletitle{Attention is all you need}.
\newblock \bibinfo{journal}{\emph{arXiv preprint arXiv:1706.03762}}
  (\bibinfo{year}{2017}).
\newblock


\bibitem[\protect\citeauthoryear{White, Vendome, Linares-V\'{a}squez, and
  Poshyvanyk}{White et~al\mbox{.}}{2015}]%
        {White2015}
\bibfield{author}{\bibinfo{person}{Martin White}, \bibinfo{person}{Christopher
  Vendome}, \bibinfo{person}{Mario Linares-V\'{a}squez}, {and}
  \bibinfo{person}{Denys Poshyvanyk}.} \bibinfo{year}{2015}\natexlab{}.
\newblock \showarticletitle{Toward Deep Learning Software Repositories}. In
  \bibinfo{booktitle}{\emph{Proceedings of the 12th Working Conference on
  Mining Software Repositories}} \emph{(\bibinfo{series}{MSR '15})}.
  \bibinfo{publisher}{IEEE Press}, \bibinfo{address}{Piscataway, NJ, USA},
  \bibinfo{pages}{334--345}.
\newblock
\showISBNx{978-0-7695-5594-2}
\urldef\tempurl%
\url{http://dl.acm.org/citation.cfm?id=2820518.2820559}
\showURL{%
\tempurl}


\bibitem[\protect\citeauthoryear{Yang, Dai, Yang, Carbonell, Salakhutdinov, and
  Le}{Yang et~al\mbox{.}}{2019}]%
        {yang2019}
\bibfield{author}{\bibinfo{person}{Zhilin Yang}, \bibinfo{person}{Zihing Dai},
  \bibinfo{person}{Yiming Yang}, \bibinfo{person}{Jaime Carbonell},
  \bibinfo{person}{Ruslan Salakhutdinov}, {and} \bibinfo{person}{Quoc~V. Le}.}
  \bibinfo{year}{2019}\natexlab{}.
\newblock \showarticletitle{XLNet: Generalized Autoregressive Pretraining for
  Language Understanding}.
\newblock \bibinfo{journal}{\emph{arXiv preprint arXiv:1906.08237}}
  (\bibinfo{year}{2019}).
\newblock


\bibitem[\protect\citeauthoryear{{Zhang}, {Wang}, {Zhang}, {Sun}, {Wang}, and
  {Liu}}{{Zhang} et~al\mbox{.}}{2019}]%
        {8812062}
\bibfield{author}{\bibinfo{person}{J. {Zhang}}, \bibinfo{person}{X. {Wang}},
  \bibinfo{person}{H. {Zhang}}, \bibinfo{person}{H. {Sun}}, \bibinfo{person}{K.
  {Wang}}, {and} \bibinfo{person}{X. {Liu}}.} \bibinfo{year}{2019}\natexlab{}.
\newblock \showarticletitle{A Novel Neural Source Code Representation Based on
  Abstract Syntax Tree}. In \bibinfo{booktitle}{\emph{2019 IEEE/ACM 41st
  International Conference on Software Engineering (ICSE)}}.
  \bibinfo{pages}{783--794}.
\newblock
\showISSN{0270-5257}
\urldef\tempurl%
\url{https://doi.org/10.1109/ICSE.2019.00086}
\showDOI{\tempurl}


\end{thebibliography}

\end{document}